\documentclass{article}
\usepackage{arxiv}
\usepackage[utf8]{inputenc} 
\usepackage[T1]{fontenc}    
\usepackage{hyperref}       
\usepackage{url}            
\usepackage{booktabs}       
\usepackage{amsfonts}       
\usepackage{nicefrac}       
\usepackage{microtype}      
\usepackage{lipsum}		
\usepackage{graphicx}
\usepackage{natbib}
\usepackage{doi}
\usepackage{float}

\title{XGBoost "is all you need": the case of forecasting transmitted heat energy in District Heating Systems}

\date{} 					

\author{ 
    \href{https://orcid.org/0000-0002-4141-2058}
        {\includegraphics[scale=0.06]{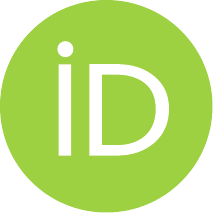}\hspace{1mm}Milan Zdravković}
            \thanks{This research was supported by the Science Fund of the Republic of Serbia, Grant No. 23-SSF-PRISMA-206, Explainable AI-assisted operations in district heating systems - XAI4HEAT} \\
	Faculty of Mechanical Engineering\\
	University of Niš\\
	ul. Aleksandra Medvedeva 14, 18000 Niš, Serbia \\
	\texttt{milan.zdravkovic@masfak.ni.ac.rs} \\
}

\renewcommand{\headeright}{}
\renewcommand{\undertitle}{}

\hypersetup{
pdftitle={A template for the arxiv style},
pdfsubject={q-bio.NC, q-bio.QM},
pdfauthor={David S.~Hippocampus, Elias D.~Striatum},
pdfkeywords={First keyword, Second keyword, More},
}

\begin{document}
\maketitle
\begin{abstract}
	This paper presents a comparative study of two distinct approaches, XGBoost and Long-Short Term Memory (LSTM), for forecasting transmitted heat energy in District Heating Systems (DHS). The objective is to explore scenarios in which conventional ML algorithms demonstrate better performance over deep learning networks in time series forecasting and the associated benefits in terms of computational cost and environmental impact. The study focuses on a real-world DHS dataset. Through experimentation and analysis, it is demonstrated that XGBoost consistently outperforms LSTM in this specific forecasting task. The difference is explained by the error distribution illustrating that LSTM makes more significant errors in the intervals of less data availability. The reduced computational demands of conventional ML approaches not only result in cost savings but also minimize the carbon footprint associated with data analysis tasks in energy systems.
\end{abstract}

\bigskip
\noindent\textbf{Publication note.}
This preprint corresponds to the paper published as:

Zdravković, M. (2024). XGBoost “is All You Need”: the Case of Forecasting Transmitted Heat Energy in District Heating Systems. In: Trajanović, M., Filipović, N., Zdravković, M. (eds) Disruptive Information Technologies for a Smart Society. ICIST 2024. Lecture Notes in Networks and Systems, vol 860. Springer, Cham. 
The final authenticated version is available at
\url{https://doi.org/10.1007/978-3-031-71419-1_2}.

\section{Introduction}
The extreme popularity and availability of off-the-shelf deep learning algorithms and architectures today have created excitement and very high expectations related to quickly addressing different automation challenges in different industries. The promise of simplicity of use combined with performance already demonstrated mostly in the cases of language processing and computer vision, has led to a surge in applying these technologies across various sectors. These include healthcare, finance, automotive, and more, where they are being used for tasks like disease detection, financial forecasting, autonomous driving, and customer service automation. However, this enthusiasm must be tempered with a recognition of the complexities involved. Successful implementation often requires large amounts of high-quality data, and the ability to interpret and fine-tune models to specific needs. Moreover, issues like algorithmic bias, transparency, and ethical considerations pose additional challenges.
The reality is that while deep learning offers powerful tools, their effective application demands more data, more care and more expertise. Quite often, Deep Learning architectures are tested quickly and applied without careful consideration and with prejudice driven by AI hype. This hasty adoption often leads to overlooking crucial aspects like algorithm suitability, data quality, and computational requirements. The result is systems that either underperform or consume excessive energy, thus negating the benefits of using AI. This situation underscores the importance of a more measured approach to implementing Machine Learning solutions, one that involves thorough testing, consideration of environmental impact, and an understanding of the specific problem context. Moreover, it emphasizes the need for organizations to invest in building or acquiring the necessary expertise to harness the full potential of AI technologies effectively and sustainably.

The objective of research behind this paper is to demonstrate that traditional ML algorithms are indeed competitive when compared to complex neural network algorithms in certain time series forecasting problems. This will be showcased on the example of forecasting transmitted heat energy in District Heating Systems.

Despite the maturity of District Heating systems (DHS), substantial opportunities exist for enhancing their operational efficiency. This particularly pertains to the reduction of fuel consumption costs and minimization of carbon emissions. A promising strategy involves capitalizing on the considerable potential for re-engineering current short- and long-term operational strategies of DHS. This can be achieved through the utilization of precise heat demand forecasts. Such forecasts are crucial for optimizing heat production, which leads to reduced fuel consumption, waste, and CO2 emissions. Additionally, this approach ensures maximum consumer satisfaction and facilitates more effective planning for both short and long-term heat production.
The forecasting problem and methodology are described in Section 2. Section 3 introduces and describes two competitive approaches, and it presents the implementation of experiments with the two proposed and argued architectures. Section 4 discusses the final results.

\section{Methodology}
\label{sec:headings}

In essence, the operation of District Heating System (DHS) plants involves either automated or semi-automated management of primary (at the plant level) and secondary (at the substations level) supply water temperatures, as well as water flow in the primary supply. The primary and secondary flows are closed loop and the energy from primary to secondary lines is exchanged through a heat exchanger. The management of supply water temperature is based on the collective DHS demand and prevailing weather conditions. The overall DHS demand is determined by the transmitted heat energy in the specified interval, namely the difference between the measurements of transmitted energy in the current and past timepoint, recorded at the calorimeter, located at the return primary line.

Presently, conventional DHSs are managed through a Supervisory Control And Data Acquisition (SCADA) system. This system integrates various sensors, control mechanisms, and algorithms that automatically modify operational parameters in response to sensor data. DHS control at the district heating substation levels (primary and secondary sides of DHS) is automated. This includes the implementation of appropriate hot water reset controls (outdoor air reset or control curve), often de-scribed as a regulation curve, used by SCADA system to deduce desired supply line water temperature based on the measured outside air temperature.
The process is described in detail and forecasting model elaboration is provided in the earlier work \cite{Zdravkovic2022}. This paper examines the potential to replace simplistic control curve with a model capable to forecast the transmitted energy based on the measured air temperature in the previous timepoint.

The selected comparative methods are stacked Long-Short Term Memory architecture and Gradient Boosting approach, namely its XGBoost implementation. The experiment involves preparing the data appropriately for each model, training both models, and then evaluating and comparing their performance using relevant metrics. Additionally, XGBoost model is optimized by finding the set of hyperparameters providing the best metrics. The method used was Bayesian optimization.

The metrics used were Root Mean Square Error (RMSE), Mean Absolute Error (MAE) and Coefficient of determination (R2 score). The coefficient of determination, often denoted as R2 (R-squared) is a statistical measure that represents the pro-portion of the variance for a dependent variable that's explained by an independent variable or variables in a regression model. Additional metrics for comparison are time for training and inference on the test set. The experiment is carried out in Google Colab environment, using T4 GPU runtime.

Implementation of XGBoost for regression problems expects a structured dataset, not a time series. Transforming time series data into a format suitable for regression problems is a common approach in Machine Learning for forecasting and other time-dependent analyses. This process involves converting the sequential nature of time series data into a structured format that a regression model can understand. One standard method is to create lagged features, which are values from previous time steps used as separate input features. The number of lagged features (also known as the lag order) depends on the specific problem and how far back in the past the predictive patterns extend. Additionally, certain qualities of the times of the measured instances will be extracted and used as features, namely, hour of the day, day of the week and month.

\section{Implementation and discussion}

Data from one substation, namely substation 9 from the local DHS, was used for the experiment. Data included outside ambient temperature from the sensor located at the building facade, on the north side; temperatures of water in supply and return primary and secondary lines and transmitted energy, measured by the calorimeter located at the primary return line. Two heating seasons were considered for analysis, namely 2018/19 and 2019/20, in total 5832 timepoints.

Ambient temperature and transmitted energy were used in model training, while the other features were dropped. Transmitted energy in the selected timepoint is calculated as the difference of the energy reading in that timepoint and in the previous one. Only the period from November to March was considered in the analysis. Normally, heating season starts mid-October and lasts till mid-April. However, those periods are characterized by high variance in temperature, special regimes of operation and thus, will not be accounted for. The missing data was found in the dataset. This was due to the lack of 3G network connectivity at certain time points. Missing data were imputed by using linear interpolation.

Overall distribution of available data is displayed in figure 1, after resetting date time index in order to enable a continuous signal presentation. Also, zoomed in overlay plot illustrates the correlation between the outside temperature and transmit-ted heat energy.

\begin{figure}[H]
	\centering
	\includegraphics[width=1\textwidth]{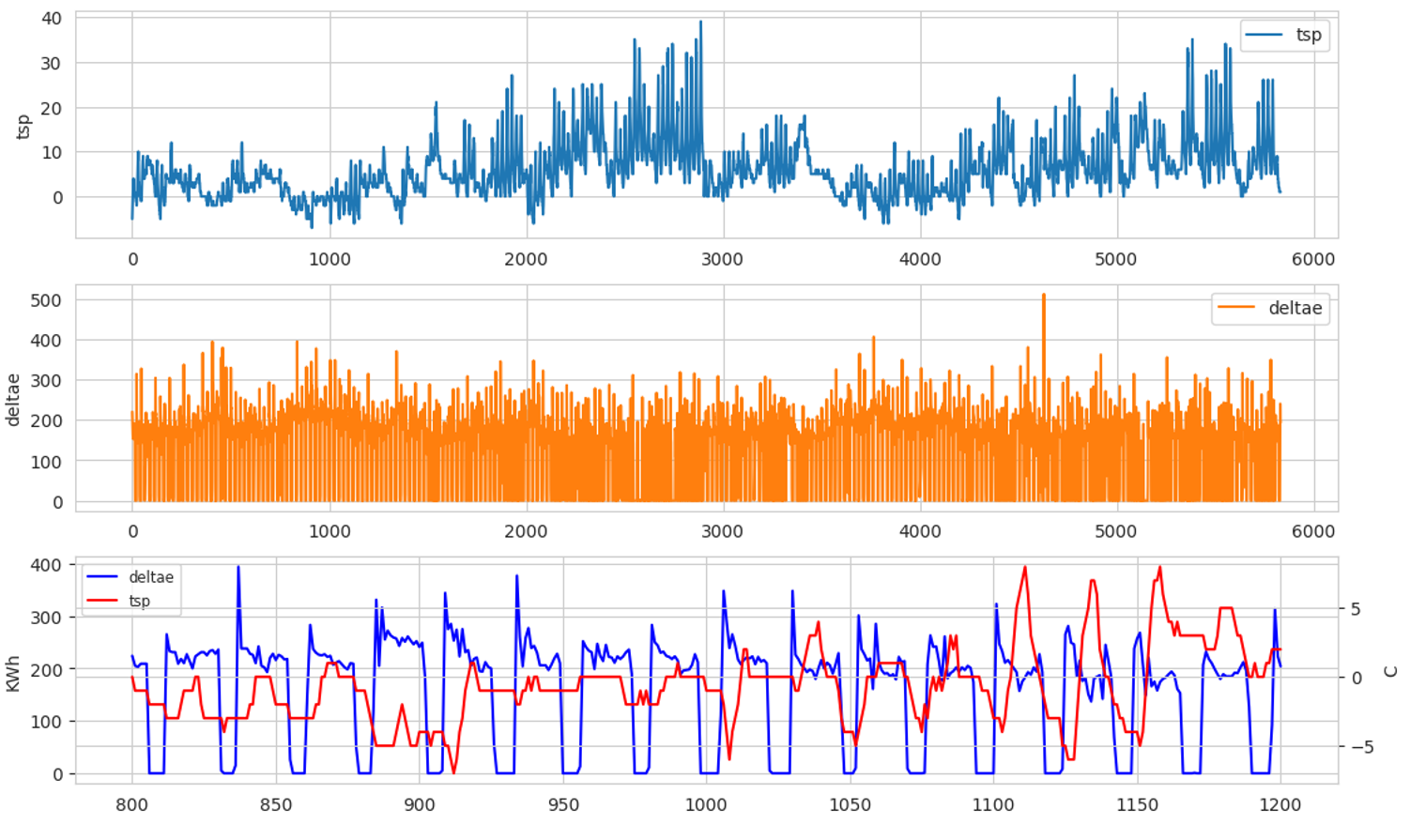}
	\caption{Ambient temperature (top), transmitted energy time series data (mid) and zoomed-in overlay of two signals.}
	\label{fig:fig1}
\end{figure}

Both time series are stationary, which is confirmed by the Augmented Dickey-Fuller (ADF) test. In a stationary time series, the mean, variance, and autocorrelation structure remain constant across different time points. Stationarity is an important requirement for the good performance of parametric algorithms, such as neural networks.

The distribution analysis of two relevant time series signals (distribution of data points - histogram, and the underlying probability density - Kernel Density Estimate plot are presented in a figure 2) indicate relatively high sparsity of hourly transmit-ted energy feature, where zeros in certain time points indicate that the system is not operational, mostly in cases of high outside temperatures. When this is ignored, deltae signal exhibits normal distribution. Normal distribution is also exhibited by the outside temperature signal, with minor skewness towards higher temperatures.

The value of Spearman coefficient (-0.308) and p-value (0.000) indicate that there is statistically significant negative association between two signals, which is expected. Even though both signals are normally distributed, Spearman values are considered instead of Pearson coefficients as more reliable indicator of association because of relatively high sparsity of deltae signal and zero data which can be also interpreted as outliers to which Spearman coefficient exhibits better response. Be-sides less sensitivity to outliers, the Spearman coefficient indicates monotonic relationship and does not assume linear association.

\begin{figure}[H]
	\centering
	\includegraphics[width=1\textwidth]{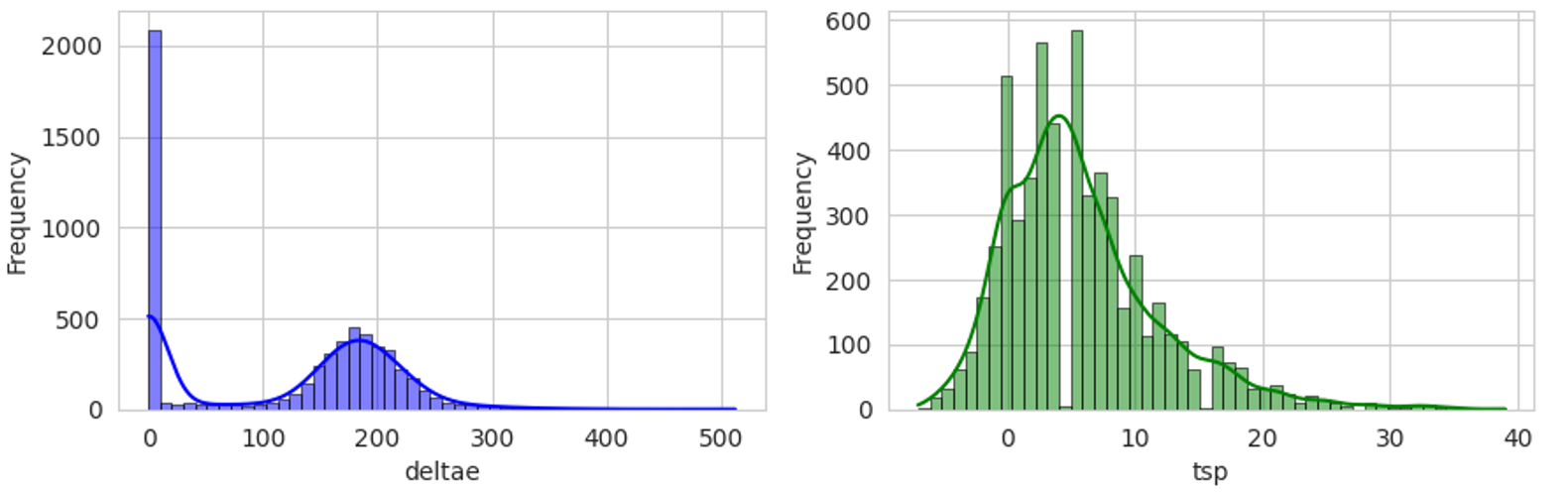}
	\caption{Distribution of transmitted energy (left) and ambient temperature (right) with KDE line.}
	\label{fig:fig2}
\end{figure}

Two models and their respective forecasting capabilities will be tested with the data above, namely stacked LSTM model and XGBoost.

\subsection{Implementation of stacked LSTM model}

Long Short-Term Memory (LSTM) networks \cite{Hochreiter1997} are a special kind of Recurrent Neural Network (RNN) \cite{Rumelhart1985}, specifically designed to learn from sequences of data and remember long-term dependencies in the data. They are widely used for sequence prediction problems, such as time series forecasting, natural language processing, and speech recognition. LSTMs are design to address the specific limitation of traditional RNNs related to struggling to capture long-term dependencies in a sequence due to the vanishing gradient problem.

\begin{figure}[H]
	\centering
	\includegraphics[width=0.8\textwidth]{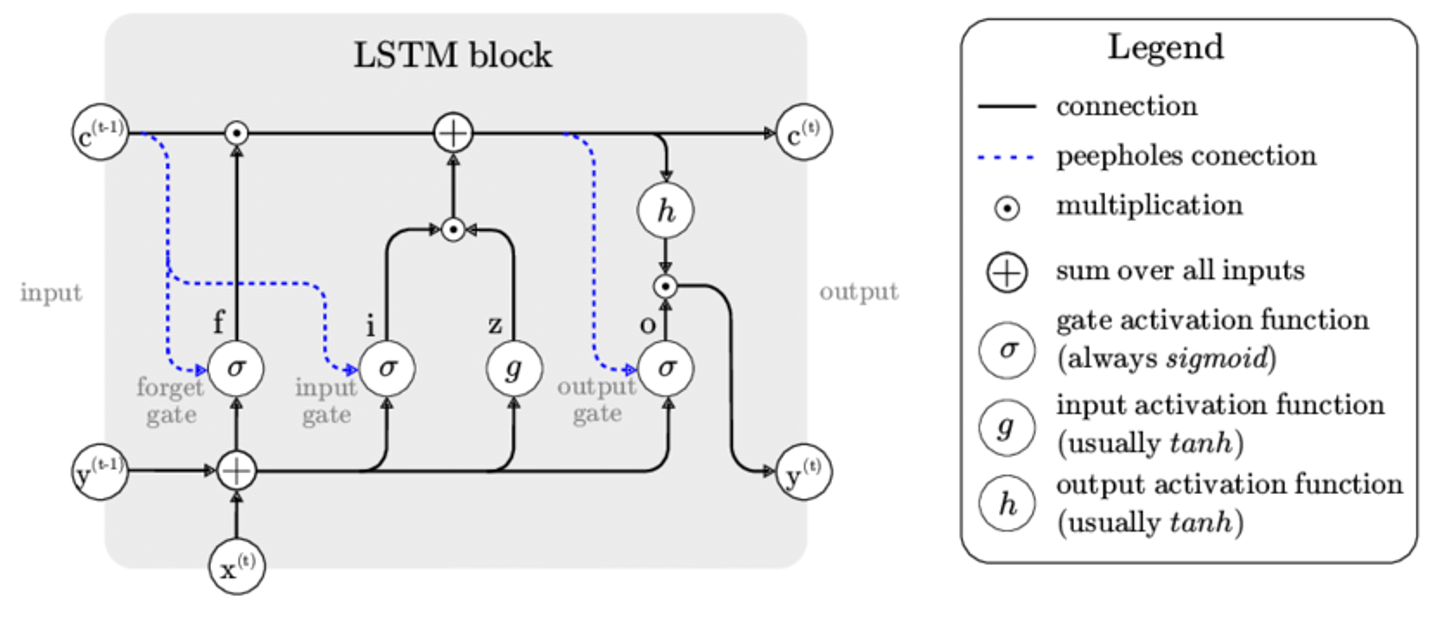}
	\caption{Architecture of a typical vanilla LSTM block \cite{VanHoudt2020}.}
	\label{fig:fig3}
\end{figure}

LSTMs maintain a hidden state vector and a cell state vector across time steps, which help them store and manage long-term dependencies in the data. An LSTM unit consists of three main components: forget gate which "decides" what information should be thrown away or kept, input gate that updates the cell state, and output gate which "decides" what the next hidden state should be. In each time step, the LSTM cell takes three pieces of information: the current input data, the previous hidden state, and the previous cell state. Based on these inputs, it produces a new hidden state and a new cell state, which are passed to the next time step (see Figure 3).

LSTM networks take structured data transposed to supervised regression problem format by introducing certain number of datapoints from the past in one instance, namely lagged timepoints. For both experiments, 6 timepoints in the past will be considered in each data instance. 80\% of all data will be used for training, while 20\% will be set aside for testing the trained model.

For the case of training LSTM architecture, data was normalized. Normalization is a crucial step for pre-processing data for training neural networks, as it improves gradient descent efficiency and effectiveness by preventing local minima.
The architecture used in the experiment was stacked LSTM, with two LSTM layers each with 100 units and Rectified Linear Unit activation, each followed with dropout layer. Mean absolute error was used as a loss function and efficient Adam optimizer has been used. Training was carried out with 100 epochs and batch size of 24, where 20\% of the training set was used for validation.

\begin{figure}[H]
	\centering
	\includegraphics[width=0.6\textwidth]{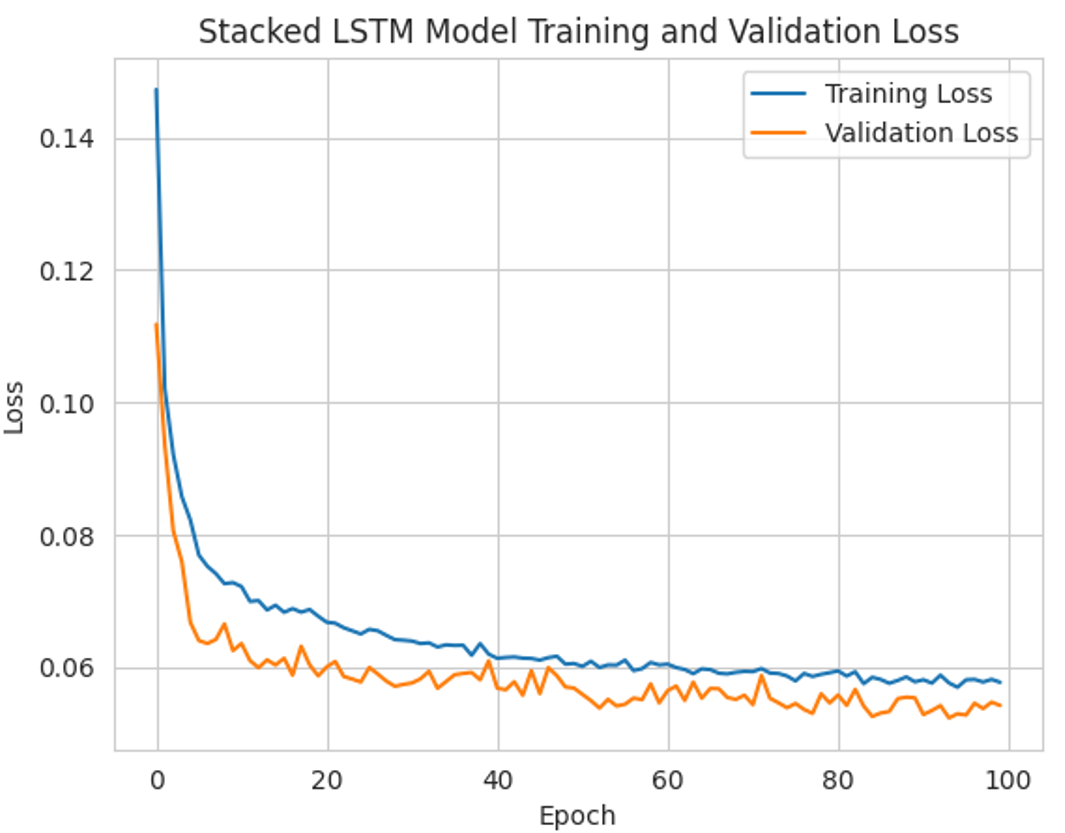}
	\caption{Stacked LSTM model training and validation loss.}
	\label{fig:fig4}
\end{figure}

Model training and validation loss curve (see Figure 4) demonstrated good generalization and it did not show overfitting, already mitigated by using dropout regularization in the model architecture.

\subsection{Implementation of XGBoost model}

XGBoost (Extreme Gradient Boosting) \cite{Chen2016} is a highly efficient and flexible algorithm widely used for supervised learning tasks. It is an implementation of gradient boosted decision trees designed for speed and performance. XGBoost has gained popularity in machine learning competitions and practical applications due to its effectiveness and efficiency in handling various types of tabular data and relevant tasks.

XGBoost is an ensemble learning method, specifically a boosting technique. It builds the model in stages, and each stage adds new models to correct the errors made by the existing ensemble of models. It operates within the gradient boosting framework \cite{Breiman1997} by constructing a new model that adds to an existing ensemble of models in a way that minimizes the overall prediction error. The "gradient boosting" part refers to the algorithm's use of the gradient descent algorithm to minimize the loss when adding new models. XGBoost primarily uses decision trees as its base learners. Each new tree corrects the residual errors (differences between predicted and actual values) of the previous trees. A key feature that differentiates XGBoost from other gradient boosting methods is its built-in regularization (both L1 and L2), which helps to prevent overfitting and improve model generalization. XGBoost can automatically handle missing data, making it robust to problems with incomplete datasets. It is optimized to efficiently handle sparse data. It incorporates a learning rate (shrinkage), which scales the contribution of each new tree added to the model. This can be used to prevent overfitting.

Feature engineering practices ensure that time dimensions are accurately reflect-ed in the relevant features. Given the cyclic nature of district heating systems operation, it is assumed that hour of day is one of the significant features. Besides that, relevance of day of the week is expected to be non-trivial, especially when considering if the day is a working day or not. Spearman correlation coefficients show statistically significant association between hour of the day and month with transmitted heat energy (SPhour=-0.339, SPmon=-0.150). Interestingly, the hypothesis on the association between day of the week and transmitted heat energy was not confirmed.

For optimizing hyperparameters of XGBoost regressor, a Bayesian approach was used. Bayesian optimization is an optimization strategy, particularly useful for Machine Learning scenarios where the evaluation of the objective function (such as model validation loss) is computationally expensive. Bayesian optimization is a probabilistic model-based approach. It constructs a posterior distribution of functions (probability model) that best describes the function that needs optimization, based on past evaluations. The process is iterative, and it starts with a set of initial hyperparameter combinations (often chosen randomly). Then, modeling the objective function is carried out, by using the results from initial and ongoing evaluations.

In the optimization case, MAE was used as the objective function. The space of hyperparameters was defined. For implementation of the Bayesian optimization, Hyperopt \cite{Bergstra2015} package was used. It is an open-source Python library used for optimizing the hyperparameters of machine learning algorithms. The Tree of Parzen Estimators (TPE) was used as an optimization algorithm. The optimization process was carried out in 80 iterations.

\section{Discussion of results}

XGBoost model outperformed LSTM model in all metrics (see summary results in Table 1). Optimization somewhat improved the performance of the model with default set of hyperparameters. In both cases, RMSE was almost double the Mean Absolute Error (MAE). RMSE gives more weight to larger errors due to the squaring of each error before averaging, while MAE treats all errors equally. RMSE being significantly higher than MAE suggests wider spread of errors or the presence of some large errors in predictions. In general, the model shows good accuracy, but it makes a few substantial errors.

\begin{table}[H]
\centering
\caption{Summary results of experiment}
\label{tab:model_comparison}
\begin{tabular}{lccc}
\toprule
\textbf{Metric} & \textbf{LSTM} & \textbf{XGB} & \textbf{XGB Optimized} \\
\midrule
RMSE (kW)                  & 62.111 & 36.700 & 35.677 \\
MAE (kW)                   & 28.890 & 20.589 & 18.687 \\
$R^2$                      & 0.540  & 0.853  & 0.861 \\
Training time (s)          & 476.129 & 0.238 & 549.237 \\
Inference time (s)         & 0.537  & 0.007  & 0.035 \\
Training CO$_2$ (g)        & 9.245  & 0.005  & 10.664 \\
Inference CO$_2$ (g/1000)  & 10.421 & 0.141  & 0.676 \\
\bottomrule
\end{tabular}
\end{table}

The histograms displayed at figure 5 show the distribution of errors (the differences between predicted and actual values) in the case of optimized XGBoost and LSTM approach. The red dashed line at zero visualize the point where there is no error (perfect prediction). The Kernel Density Estimate (KDE) line provides a smooth curve representing the error density.

\begin{figure}[H]
	\centering
	\includegraphics[width=0.8\textwidth]{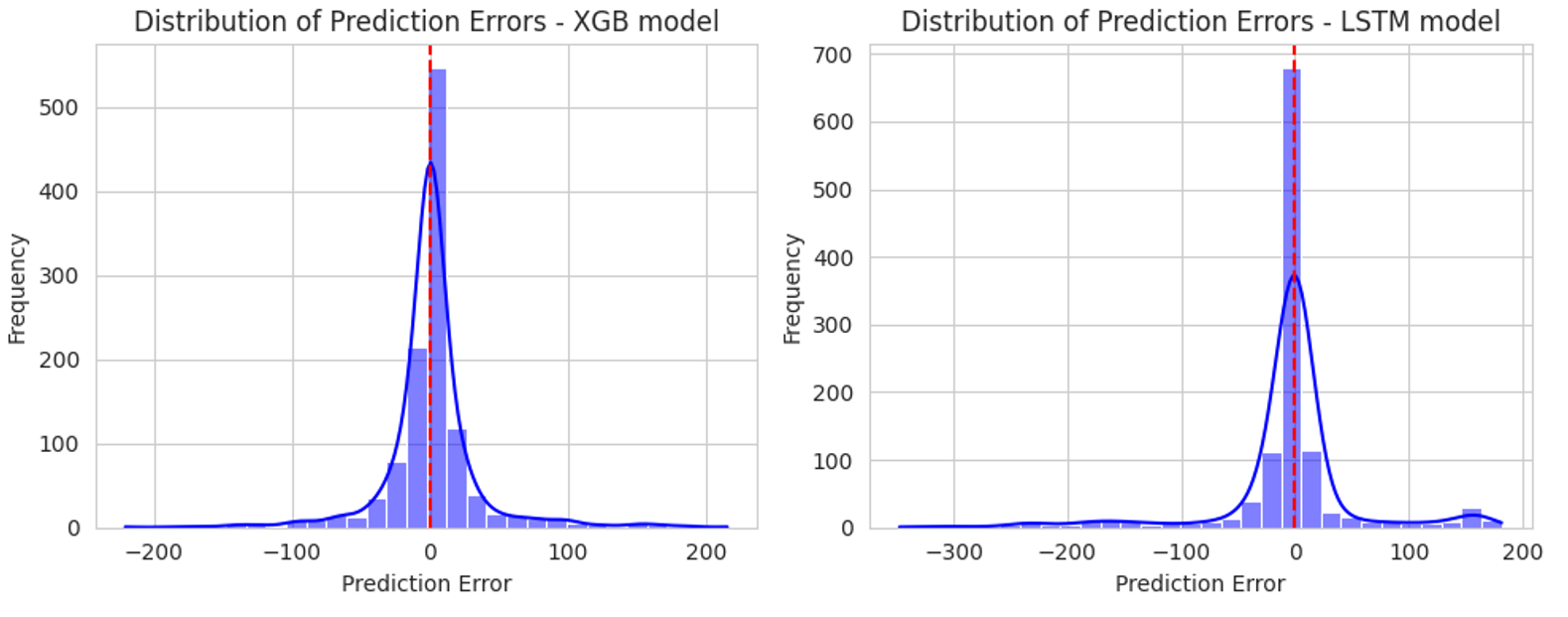}
	\caption{Distributions of prediction errors by XGBoost (left) and stacked LSTM model (right).}
	\label{fig:fig5}
\end{figure}

Error distribution histogram confirmed the assumption on good accuracy with few substantial errors in using both approaches made earlier. However, it also explained the difference in performance of both methods. Occurrence of minor fat tails - increased accumulation of larger errors in case of LSTM models explains the source of difference in performance metrics: while LSTM model has more successful forecasts with less error, it also makes more substantial errors that affect the MAE and especially RMSE.

The figure 6 shows the scatter plot of actual and predicted values for the optimized XGBoost model and LSTM model, as well as overlayed distribution of test data.

\begin{figure}[H]
	\centering
	\includegraphics[width=0.8\textwidth]{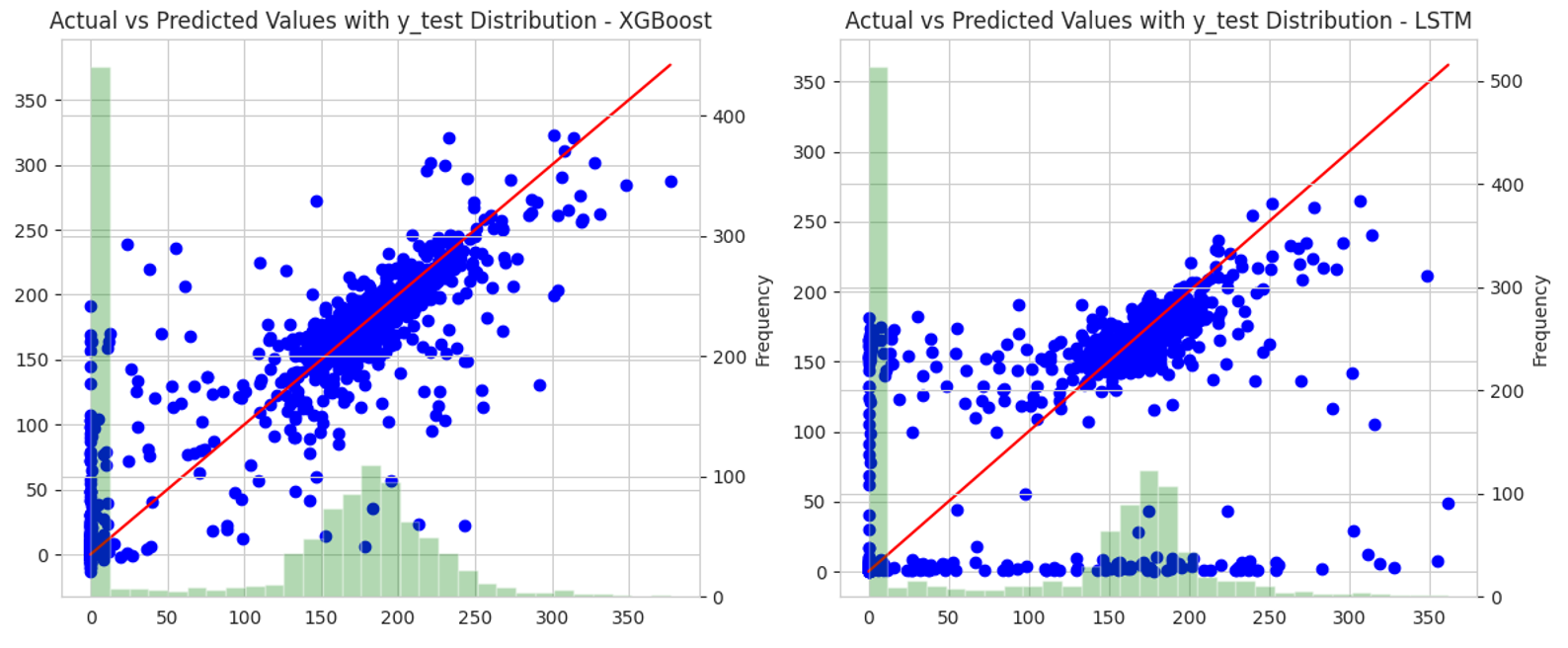}
	\caption{Plot of actual and predicted values with overlayed distribution of test data.}
	\label{fig:fig6}
\end{figure}

The figure illustrates the accuracy of the model in the different intervals of test data. In both cases, the scatter plot points are clustered along the red line, which suggests the predictions are reasonably good. However, there is some variance in both models (more significant in LSTM than XGBoost), especially for lower and higher values where the points tend to diverge more from the line. This variance can be explained by lack of data in intervals of lower and higher values for transmitted energy.

Additionally, scatter plots unveil the specific nature of significant errors made by both models, mostly due to impossibility to forecast the periods in which the heating is turned off by the operator (for XGBoost and LSTM models, concentration of fore-casts along Y-axis) or to confusing those periods with periods in which the heating is actually on (LSTM models, concentration of forecasts along X-axis). It’s important to emphasize that the decision to turn on the heating system lies with a human operator, and it is affected by the reasons not included as features in this dataset.

In general, the LSTM models are expected to show good performance at uncovering very complex patterns of heat demand that occur at the beginning and end of the heating season, when heat demand, as well as outside temperature exhibit high variance. However, there is no sufficient data to unleash the power of LSTM's long-term memory in the case of this experiment.

Furthermore, neural networks require data imputation. In this case, occasional missing data is replaced by using simplistic linear interpolation technique. It’s worth highlighting that many of the data imputation approaches (besides stochastic ones) introduce regularities that are considered as bias that can lead to better results than in reality. XGBoost assumes initial transformation of sequential time series data to structured, tabular data suitable for traditional regression problems. This appears very useful in industrial application where the periods of missing data due to sensor faults are frequent. Such faults and corresponding missing data problems cannot be addressed with imputation techniques, since those periods can be quite long. Ignoring sequential nature of data dramatically improves usability of data islands, occur-ring in such circumstances.

What are the possible reasons for XGBoost outperforming LSTM architecture in this case? First of the reasons is the size of the dataset combined with the time feature engineering practices. XGBoost can benefit significantly from good feature engineering if those features encapsulate the temporal dynamics well. LSTM is in-deed very good in uncovering these dynamics, but only if there is sufficient data available. Second reason is that LSTMs can be particularly sensitive to the choice of hyperparameters, especially when considering the network topology. The optimization has not been done in this case as it would require significant computational resources and time.
Besides accuracy, computational requirements for the two methods are not com-parable, especially when inference is considered. Training times in both cases appear to be similar, but only when optimization of hyperparameters is involved in XGBoost case. The amount of energy used for training a neural network on a T4 GPU depends on several factors, including network architecture complexity, training dataset size, batch size, choice of optimizer and learning rate. Based on the technical specifications of the manufacturer  the power usage of NVidia T4 GPU is 70 watts, corresponding to the consumption of energy of 0.07 kWh. This accounts for low-power scenarios: for small, simple networks with small datasets and low batch sizes.

The average emission factor for CO2 emission for thermal power plants is 0.998 kg/kWh \cite{Chowdhury2004}, although the actual emissions may vary depending on the specific fuel mix and efficiency of the power plants used to generate the electricity.  Assuming the energy is produced by thermal plants, CO2 emission per unit time of ML training is 0.0699 kg/h. This calculation considers large simplifications, where the most significant are assumptions on: 1) constant GPU's consumption of energy, 2) same consumption of GPU and CPU, where latter is used for training of XGBoost model, 3) constant availability of resources making comparisons based on the times needed for training and inference questionable. However, for this purpose the results interpreted by adopting those assumptions provide a good enough indication when considering the convincing advantage that could be gained by appropriately choosing the approaches to solve time series forecasting problems, as presented in the table above.

Finally, another advantage of conventional ML models is that they offer out-of-the-box functionality facilitating the interpretability of the models. The figure above shows the feature importances (see Figure 7), as interpreted by the trained XGBoost model with optimal hyperparameters. The feature importance is typically represented by a metric called "F-score." The F-score is a measure of feature importance based on how often a feature is used to split data across all the trees in the ensemble (the boosted trees in the XGBoost model). F-scores demonstrate the highest importance of the transmitted energy in the current hour, current ambient temperature and hour of the day.

\begin{figure}[H]
	\centering
	\includegraphics[width=0.6\textwidth]{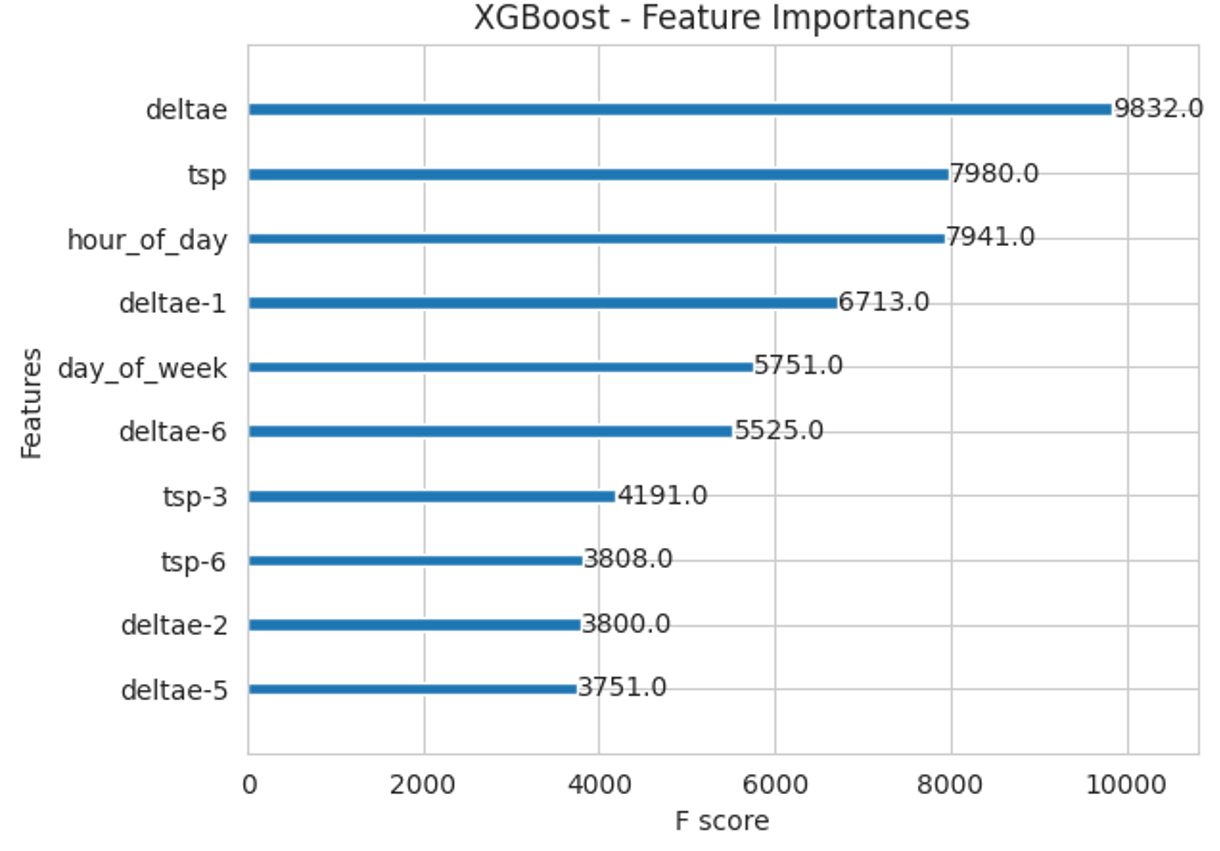}
	\caption{Feature importances in the XGBoost model.}
	\label{fig:fig7}
\end{figure}

\section{Conclusion}

The challenges of high computational cost of the modern "off-the-shelf" deep learning architectures can be often addressed by considering the "cheaper" alternatives in very efficient conventional Machine Learning approaches. Indeed, such approaches must be the first choice in case of relatively smaller datasets and sometimes - datasets with sparse and/or missing data. On such sets, in most of the cases, conventional ML algorithms will not only appear as more sustainable in context of computational requirements but also more accurate. The necessary condition for this better performance is feature engineering practices that reflect the domain experience, expert knowledge on the internal and external factors affecting the time series data and expectations related to the behavior of the systems producing the data.

Besides accuracy and sustainability, where trained deep learning architectures are often considered as "black-box" models, standard ML algorithms offer much better interpretability which is sometimes crucial in industrial applications. This interpretability is inherent and feature importances are easy to show.

\bibliographystyle{unsrt}
\bibliography{references}

@article{Zdravkovic2022,
  author  = {Milan Zdravkovi\'c and Marko Ignjatovi\'c and Ivan \'Ciri\'c},
  title   = {Explainable Heat Demand Forecasting for the Novel Control Strategies of District Heating Systems},
  journal = {Annual Reviews in Control},
  volume  = {53},
  pages   = {405--413},
  year    = {2022},
  doi     = {10.1016/j.arcontrol.2022.03.009}
}

@article{Hochreiter1997,
  author  = {Sepp Hochreiter and J{\"u}rgen Schmidhuber},
  title   = {Long Short-Term Memory},
  journal = {Neural Computation},
  volume  = {9},
  number  = {8},
  pages   = {1735--1780},
  year    = {1997},
  doi     = {10.1162/neco.1997.9.8.1735}
}

@techreport{Rumelhart1985,
  author      = {David E. Rumelhart and Geoffrey E. Hinton and Ronald J. Williams},
  title       = {Learning Internal Representations by Error Propagation},
  institution = {Institute for Cognitive Science, University of California},
  address     = {San Diego, CA},
  number      = {ICS 8504},
  year        = {1985}
}

@article{VanHoudt2020,
  author  = {Gilles Van Houdt and Carlos Mosquera and Gonzalo N\'apoles},
  title   = {A Review on the Long Short-Term Memory Model},
  journal = {Artificial Intelligence Review},
  volume  = {53},
  pages   = {5929--5955},
  year    = {2020},
  doi     = {10.1007/s10462-020-09838-1}
}

@inproceedings{Chen2016,
  author    = {Tianqi Chen and Carlos Guestrin},
  title     = {XGBoost: A Scalable Tree Boosting System},
  booktitle = {Proceedings of the 22nd ACM SIGKDD International Conference on Knowledge Discovery and Data Mining (KDD '16)},
  pages     = {785--794},
  year      = {2016},
  publisher = {ACM},
  doi       = {10.1145/2939672.2939785}
}

@techreport{Breiman1997,
  author      = {Leo Breiman},
  title       = {Arcing the Edge},
  institution = {Department of Statistics, University of California, Berkeley},
  number      = {Technical Report 486},
  year        = {1997}
}

@article{Bergstra2015,
  author  = {James Bergstra and Brent Komer and Chris Eliasmith and Daniel Yamins and David D. Cox},
  title   = {Hyperopt: A Python Library for Model Selection and Hyperparameter Optimization},
  journal = {Computational Science \& Discovery},
  volume  = {8},
  number  = {1},
  pages   = {014008},
  year    = {2015},
  doi     = {10.1088/1749-4699/8/1/014008}
}

@inproceedings{Chowdhury2004,
  author    = {S. Chowdhury and S. Chakraborty and S. Bhattacharya and A. Garg and A. P. Mitra and I. Mukherjee and N. Chakraborty},
  title     = {An Emission Estimation of Greenhouse Gas Emission from Thermal Power Plants in India During 2002--03},
  booktitle = {Proceedings of the Workshop on Uncertainty Reduction in Greenhouse Gas Inventories},
  editor    = {A. P. Mitra and Subodh K. Sharma and S. Bhattacharya and A. Garg},
  publisher = {Ministry of Environment and Forests, Government of India},
  address   = {New Delhi, India},
  pages     = {16--22},
  year      = {2004}
}

\end{document}